\documentclass[twoside]{article}
\usepackage{PRIMEarxiv}
\usepackage[utf8]{inputenc} 
\usepackage[T1]{fontenc}    
\usepackage{hyperref}       
\usepackage{url}            
\usepackage{booktabs}       
\usepackage{amsfonts}       
\usepackage{nicefrac}       
\usepackage{microtype}      
\usepackage{lipsum}
\usepackage{listings}
\usepackage{fancyhdr}       
\usepackage{graphicx}       
\graphicspath{{media/}}     
\usepackage{subcaption}
\usepackage{calc}
\usepackage{amssymb}
\usepackage{amstext}
\usepackage{amsmath}
\usepackage{amsthm}
\usepackage{multicol}
\usepackage{pslatex}
\usepackage{apalike}
\usepackage{algorithm2e}
\usepackage[bottom]{footmisc}

\usepackage{multirow}
\usepackage{graphicx}
\usepackage{xcolor}
\usepackage{verbatim}

\title{Two new context-free Nahuatl grammars for automatic corpora expansion 
\thanks{\textit{\underline{Citation}}: 
\textbf{Guzman-Landa et al. Two new context-free Nahuatl grammars for automatic corpora expansion. ArXiv, 15 pp.}} 
}

\author{Juan-José Guzm\'an-Landa$^1$, Juan-Manuel Torres-Moreno$^1$, Graham Ranger$^2$\\ 
$^1$Laboratoire Informatique d'Avignon, $^2$Laboratoire Identités Culturelles, Textes et Théâtralité \\ 
Avignon Université, France\\
\texttt{\{juan-jose.guzman-landa, juan-manuel.torres, graham.ranger\}@univ-avignon.fr}
\AND
Miguel Figueroa-Saavedra$^3$, Ligia Quintana-Torres$^4$, Martha-Lorena Avenda\~no-Garrido$^4$\\ 
$^3$Instituto de Investigaciones en Educación, $^4$Facultad de Matem\'aticas\\
Universidad Veracruzana, Xalapa, Mexico\\
\texttt{\{migfigueroa, ligiaquintana, maravendano\}@uv.mx}
}

\begin{document}
\maketitle

\begin{abstract}
The aim of this article is to introduce two  Context-Free Grammars (CFG) for Nawatl Corpora expansion. Nawatl is an Amerindian language (it is a National Language of Mexico) of the $\pi$-language type, i.e. a language with few digital resources. For this reason the corpora available for the learning of Large Language Models (LLMs) are virtually non-existent, posing a significant challenge. The goal is to produce a substantial number of syntactically valid artificial Nawatl sentences and thereby to expand the corpora for the purpose of learning non contextual embeddings. For this objective, we introduce two new Nawatl CFGs and use them in generative mode. Using these grammars, it is possible to expand Nawatl corpus significantly and subsequently to use it to learn embeddings and to evaluate their relevance in a sentences semantic similarity task. The results show an improvement compared to the results obtained using only the original corpus without artificial expansion, and also demonstrate that economic embeddings often perform better than some LLMs.
\end{abstract}

\keywords{Nawatl language; Symbolic NLP; Corpus; Context-Free Grammars; Sentences semantic similarity.}

\section{Introduction}
Nahuatl or \textit{Nawatl} is one of Mexico's national indigenous languages and is recognized as the second national language after Spanish with the second highest number of speakers, at approximately 1.65 million people \cite{inegi2020censo}. This language has considerable dialectal diversity, with 29 recognized varieties, distributed across four major regions: western, central, eastern, and Huasteca (see Ethnologue, 2025\footnote{\url{https://www.ethnologue.com}} and \cite{Lastra1986areas}).
This linguistic diversity poses challenges for the use of corpora in educational, communicative, and digital contexts, as it involves significant variation in spelling and lexical selection. \cite{zimmerman,olko2016bridging,hansen2024nahuatl}.

Although the publication of digital materials and content in the Nawatl language has increased, their notable dispersion and variety has prevented them from gaining visibility on social media and from being clearly identified and accessed in repositories. They are nonetheless seeing a gradual increase in their presence and literate use, even though the available digital linguistic resources remain limited for this essential element in the current revitalization of the language \cite{pugh-etal-2025-ihquin}.

For example, only one automatic translator has been available for the Huasteca variant since 2024\footnote{See \textit{Google Translate} (\url{https://translate.google.com.mx/?hl=es&sl=nhe&tl=en&op=translate})}. As for repositories, in 2017 the National Autonomous University of Mexico (UNAM) created \textit{Axolotl}, a corpus of documents in bilingual Spanish/Nawatl versions.\footnote{The \textit{Axolotl} corpus can be retrieved from: \url{http://www.corpus.unam.mx/axolotl}}. These corpora feature mainly historical texts and so many varieties and texts produced from the 18th century to the present day are still not fully accessible or incorporated into tools or applications for general or specialized use and are not designed for contemporary native Nawatl speakers and writers.

A serious problem is, therefore, the scarcity of computational resources for this language. In particular, the corpora available for machine learning.
Our approach to mitigating this problem proposes using formal grammars  to generate artificial corpora that respect the structure of the Nawatl language.
The objective of the proposed grammars is not to model the entire written Nawatl language, a task that is virtually impossible given its complexity, but rather to generate syntactically valid sentences. These sentences can serve as the basis for the creation of large-scale synthetic corpora. Such corpora can be used to improve the training of large-scale Language Models (LLMs), both static and dynamic, in the case of so-called $\pi$-languages— that is, languages with limited digital resources \cite{these-pi,abdillahi:hal-01311495}.
Specifically, our goal is to expand the Nawatl corpus, $\pi$-\textsc{yalli}\footnote{This corpus can be retrieved from the website.: \url{https://demo-lia.univ-avignon.fr/pi-yalli}}, which has been previously used in semantic similarity tasks at the word \cite{torres2024pi} and phrase levels \cite{NAHU2,piyalliTALN}. 

The structure of the article is as follows: Section \ref{sec:nawatl} provides an overview of the Nawatl language and its grammatical features. Section \ref{sec:GNC} introduces context-free grammars. Section \ref{sec:GNC_proposed} presents two context-free microgrammars proposed for Nawatl. Section \ref{sec:artificial_corpus} describes the expanded corpus, $\pi$-yall-\textsc{ia}, using $\pi$-{\sc yalli} and generative grammars. Section \ref{sec:experiments} presents the experiments carried out with the corpus $\pi$-yall-\textsc{ia} in a semantic similarity task. Finally, Section \ref{sec:conclusions_future} concludes the article and suggests possible avenues for future research.
Appendices with examples of the grammars and resources used complete the manuscript.

\section{The Nawatl language }
\label{sec:nawatl}

Nawatl, as a Uto-Aztecan language, is an agglutinative and polysynthetic language that forms words by joining various morphemes to a verbal or noun root, thus constructing meaning.

Nawatl sentences have a basic syntactic order of {\bf verb–subject–object} (VSO), although this allows for structural flexibility that responds to the needs of speakers and contexts. Thus, orders such as VO, VS, VOS and, less frequently, SV, SVO and SOV are also found (see Table \ref{tab:EXEMP_GRAM}). Syntactic relationships between words and clauses are established through the valency of the verb and the use of connectors (particles).

\begin{table}[h!]  
  \begin{center}
\resizebox{0.49\textwidth}{!}{%
    \begin{tabular}{c|c|c}
     \toprule
    \bf Structure & \textbf{Example} & \textbf{Traduction} \\
    \midrule
   \bf VSO & Kitta tlakatl kalli & See (a) man (a) house \\ \hline
   \bf ~~VO & Kitta kalli & (He/she) sees (a) house \\ \hline
   \bf ~~VS & Kitta tlakatl & see (his) (a) man \\ \hline
   \bf VOS & Kitta kalli tlakatl
 & see (a) house (a) man \\ \hline \hline
   ~~SV & Tlakatl kitta & (A) man see (his) \\ \hline
   SVO & Tlakatl kitta kalli & (a) man see (a) house \\ \hline
   SOV & Tlakatl kalli kitta & (A) man (a) house see \\
    \bottomrule
  \end{tabular}
  }
 \caption{Examples of syntactic constructions in the Nawatl language. The most frequent structures (verbal phrases V beginning with a verb) are indicated in bold.\label{tab:EXEMP_GRAM}}
 \label{tab:corp_lit:stats-fr}
 \end{center}
\end{table}

These connectors can also be formed by groupings that establish nuances of meaning, as well as functioning as discourse connectors. Some of these words can constitute “one-word sentences,” since their morphology includes the subject and predicate, as well as information about the actants, modal, directional, and relational elements (see, for example, \cite{Launey1978introduction,floresnajera2019gramatica} and \cite{sasakidivide}).

\section{Context-Free Grammars}
\label{sec:GNC}

A Context-Free Grammar (CFG) is a type of formal grammar used to describe the syntax of formal languages, especially in the syntactic analysis of programming languages and natural languages. \cite{hopcroft2006automata}.

Formally, a context-free grammar is defined as a quadruple:

\begin{itemize}
\item $V$: a finite set of \textbf{non-terminal symbols}.
\item $\Sigma$: a finite set of \textbf{terminal symbols}, such that $V \cap \Sigma = \emptyset$.
\item $R$: a finite set of \textbf{production rules}, in the form:
  $  A \rightarrow \alpha $

where $A \in V$, $\alpha \in (V \cup \Sigma)$, and $(V \cup \Sigma)$ represents all possible strings (of length $\ge$ 0) formed with symbols of $V$ and $\Sigma$.

\item $S \in V$: the initial symbol.

\end{itemize}

In a CFG, rules are applied without considering the environment in which the non-terminal symbol appears. These grammars are used in the analysis of languages, both natural and formal, as a tool for modeling the hierarchical structure of sentences, since they allow us to describe how words and phrases are combined to form more complex expressions.

In linguistic analysis, CFGs are used to reconstruct the syntactic structure of a given sentence. In the field of computational linguistics, they are fundamental for the construction of parsers (or syntactic analyzers), components that receive a linear sequence of symbols—words in a sentence or tokens in a program—and generate from it a hierarchical structural representation according to the defined grammar.

CFGs also have applications in language documentation, where they help to formalize the structure of poorly described languages, for example by modeling the order of their constituents. They are also used in psycholinguistics, as hypotheses about how speakers internally represent syntactic structures, and in teaching, where they serve to show the hierarchy present in sentence construction explicitly. \cite{allen1995natural}. 

However, our hypothesis is that, when used in generative mode, CFGs could be useful for producing large amounts of artificial data (even if repetitive, forced, lacking in style, etc.) that are indispensable for language model learning in the case of $\pi$-languages. \cite{arxiv_G1}. 
There are grammatical features in Nawatl that are absent, such as gender, which makes the process easier for us than in other languages, and we will also limit our micro-grammars to certain grammatical persons.
The intention and intuition behind the use of two CFGs for Nawatl in this research project are in line with this pragmatic approach.

\section{Two Nawatl Context-Free Grammars}
\label{sec:GNC_proposed}

This section presents two context-free micro-grammars for Nawatl. These grammars are considered micro versions mainly because they both avoid the use of recursive production rules. Similarly, only the first three grammatical persons (I, you, he/she/it) are included, and verbs are limited to the singular form and the present tense.

\subsection{ $\mu${\sc gnaw}$\oplus$0: a first Nawatl CFG}

We summarize here the information from the first micro-grammar of nawatl, which is $\mu$\textsc{gnaw}$\oplus$0 (see Figure \ref{tab:gram0}), recently introduced by \cite{arxiv_G1}.
This approach is inspired by the grammatical frameworks developed for Indo-European languages. Given that Nawatl belongs to indigenous American language families —linguistically distant from Indo-European families— this model results in a limited and reductive representation of the language.

This grammar develops a classical model consisting of two types of phrases: the noun phrase ($N$) and the verb phrase ($V$). Both types can include additional elements (particles, nouns, and prefixes) that translate into our grammatical categories as temporal adverbs and quantifiers ($ADV_T, ADV_Q$), adjectival nouns ($ADJ$), personal pronouns ($PP$) and person markers ($PV$), possessive markers ($POS$) and negation ($NEG$).
Although derived from the grammars of families of $\tau$-languages, this preliminary version nevertheless allows for the generation of some basic structures of Nawatl.

\lstset{
   basicstyle=\footnotesize\ttfamily,
  mathescape=true
}

\label{sec:grammar}
\begin{figure}[h!]
\hrule
\begin{lstlisting}
P -> ADV$_T$ (N|V)
N -> ADJ (ART_|POS)+n 
V -> N NEG PV$_3$+v AD$V_Q$
V -> PP$_i$ NEG PV$_j$+v ADV$_Q$; i,j=1,2,3; i=j

ADV$_Q$ -> list of adverbs of quantity
ADV$_T$ -> List of adverbs of time 
ADJ -> List of adjectives
ART -> se|ni|empty          # one|the/this 
POS -> no|mo|i              # mine|yours|his, hers, this
PP$_i$ -> na|ta|ya              # I/my|you|he, she, it
PV$_j$ -> ni|ti|empty           # I/my|you|he, she, it
NEG -> amo|axkeman|empty    # no|never

n -> list of nouns
v -> list of verbs

$\oplus$ = concatenation $\emptyset$ = null $\_$ = space
\end{lstlisting}
\hrule
\caption{Nawatl micro-grammar $\mu$\textsc{gnaw}$\oplus$0.
\label{tab:gram0}}
\end{figure}

\subsection{$\mu${\sc gnaw}$\oplus$1: a more realistic Nawatl CFG}
\label{sec:grammar1}

The new CFG Nawatl grammar is based on the prototypical grammatical structures characteristic of the $VSO$ word order (and derivatives), as well as on the less frequent $SVO$ patterns described in Section \ref{sec:nawatl}.
The main objective is to accurately model the different types of verb phrases ($V$), which are frequently used, along with the relatively less common noun phrases ($S$).
These phrases can incorporate elements that we will call \textit{markers}.
Some of these markers have  agglutinative characteristics, and others are lexical particles that can indicate place, time, and intensity, playing a role similar to that of "adverbs" or “adjectives” in Indo-European languages. 

We have therefore defined the following: person marker ($MV$), object marker ($MO$), possessive marker ($POS$), temporal marker ($MT$), quantity marker for nouns ($MCS$), intensity marker for verbs ($MIV$), and finally the place marker ($ML$).
In addition to the above, there are terminal nodes that represent negations ($NEG$), adjectives ($ADJ$), nouns ($N$) and verbs ($V$).

To achieve a better approximation of Nawatl grammar, the following modifications have been introduced:

\begin{itemize}
\item The number \textbf{se} (one) and the particle \textbf{ni / in} (the) have been omitted;
\item The personal markers \textbf{na / neh} (I/my), \textbf{ta / teh} (you), and \textbf{ya, yeh} (he/she/it) have been excluded;
\item Quantifier and temporal adverbs have been reclassified as $MCS$, $MIV$ y $MT$;
\item Place markers have been introduced. ($ML$);
\item Nouns ($N$) now have two forms, one that allows them to be possessed (with a $POS$ tag) and one that does not.
\end{itemize}

In this micro-grammar (see Figure \ref{tab:TG1}), the verb assumes a central and predominant role in Nawatl syntactic constructions.
Thus, the formulation of this new grammar is based on structural patterns observed in the Nawatl spoken by Nahua speakers.

\begin{figure}[h!]
\hrule
\lstset{
   basicstyle=\footnotesize\ttfamily,
  mathescape=true
}
\begin{lstlisting}
P -> VSO | $\overrightarrow{\texttt{VO}}$ | $\overleftarrow{\texttt{VS}}$ | VOS | SV | SOV | SVO
V -> NEG MT MIV MV$_i$+MO$_j$+v;i,j=1,2,3;idifj
S -> MCS ADJ POS+n 
O -> ADJ POS+n ML

ADJ -> weyi|istak|empty    # big|white|empty
POS -> no|mo|i|empty       # my|your|his|empty
NEG -> amo|empty           # no|empty
MT -> aman|cemicac|empty   # now|forever|empty
MIV -> miyak|nochi|empty   # a lot|all|empty
MCS -> san|miakpa|empty    # only|often|empty
ML -> nikan|nepa|empty     # up|there|empty
MV$_i$ -> empty                # he/she
MO$_j$ -> ki                   # from him/her

V -> toka|itta|chihua|pia|maka|neki|...                   
   # bury|see|do|have|give|want|...
N -> siwatl|miston|elotl|xokotl|tochin|yolkatl|nakatl|... 
   # woman|cat|corn|fruit|rabbit|animal|meat|...
$\oplus$ = concatenation  $\emptyset$ = null
\end{lstlisting}

\hrule
\caption{Nawatl micro-grammar $\mu$\textsc{gnaw}$\oplus$1. Temporal marker (MT), noun quantity marker (MCS), verbal intensity marker (MIV), verbal marker (MV), object marker (MO), place marker (ML). See Appendix~\ref{sec:appendix4} for a complete list of the Prolog knowledge base.
\label{tab:TG1}}
\end{figure}

The symbols $\overrightarrow{VO}$ and $\overleftarrow{VS}$ reasonably indicate the reading direction of the structure to minimize ambiguities between the subject and the object.
For example, the sentence $\overleftarrow{VS}$:
$\overleftarrow{\textit{Kitta tlakatl}}$: A man sees (him, her, it), unlike the sentence $\overrightarrow{VO}$:
$\overrightarrow{\textit{Kitta kalli}}$: (He/she) sees a house.

Unlike micro-grammar $\mu$\textsc{gnaw}$\oplus$ 0, the micro-grammar $\mu$\textsc{gnaw}$\oplus$1 can establish more syntactic relationships between words in their text normal composition. To do this, we impose the following restrictions:
\begin{itemize}
    \item Exclusive use of base 1 verbs.\footnote{Verbs in Nawatl can have three bases: base 1 represents all verbs in the present tense, base 2 represents verbs in the past tense, and finally, base 3 represents verbs in the hypothetical future tense.}
    \item Exclusive use of transitive verbs and verb conjugation and valency in the 3rd person singular.
    \item Adjectives are part of the S and O structures, that is, they are positioned before POS.
    \item There are two markers of intensity or quantity: MIV is used to intensify the verb and MCS to quantify the noun.
\end{itemize}

All of this allows to add more features closer to real Nahuatl within the micro-grammar $\mu$\textsc{gnaw}$\oplus$1. 
\color{black}
On the other hand, rhetorical connectors $CR$ (and, but, more, however, etc.) add greater variability and richness to the generated sentences, and could be incorporated into the micro-grammar as follows: $P \rightarrow P ~ CR ~ P$.
However, due to its recursive nature, this rule will not be included in both grammars in their generative phase, but will be included as post-processing \ref{sec:refinamiento} in grammar $\mu${\sc gnaw}$\oplus1$ with a variable number of $CR$, in order to increase diversity and produce a more realistic textual output:
$P \rightarrow P ~ CR ~ P [[ ~ CR ~ P] ~ CR ~ P ...]$.

\section{$\pi$-{\sc yall-ia}, an artificial corpus of Nawatl with realistic characteristics}
\label{sec:artificial_corpus}

We have observed that the two micro-grammars $\mu$\textsc{gnaw}$\oplus$(0,1) are capable of generating a large number $\mathfrak{F}^{0,1}$ of grammatically acceptable productions.
However, although the values of $\mathfrak{F}^{0,1}$ are large, they represent only a tiny fraction of the number of sentences that could be produced using recursive grammars.
Nevertheless, such grammars are beyond the scope of this article.

But even without recursion, the number of sentences $\mathfrak{F}^{0,1}$ is still sufficient to allow for the artificial enrichment of the $\pi$-\textsc{yalli} corpus. However, a large number of these sentences are meaningless. In order to produce sentences that provide more relevant information for machine learning, certain additional restrictions must be established to generate sentences that are not only grammatically correct, but also semantically acceptable.
In fact, we prefer not to accept sentences such as:
``\textit{The big corn cob eats a lot of rabbit.}'',
or:
``\textit{The fruit walks too much}'',
because, even though they belong to the grammars $\mu$\textsc{gnaw}$\oplus$(0,1), they are not semantically realistic.\footnote{In a broader sense, even those phrases with debatable semantics could have a place, for example, in fantasy literature. For this reason, the fact that a certain number of \textit{unrealistic} phrases are generated artificially is not really a major problem.}. 
To solve the above, we introduce a semantic filter that will be explained in the section \ref{sec:filtros}. 

Another problem with combinatorial explosion is the large number of sentences generated with minimal grammatical differences, which have the same semantics, i.e., they are redundant.
Although this situation has been partially resolved by the exclusive use of the $VSO$ structure, the problem still persists.
For example, when using the knowledge base in micro-grammar $\mu$\textsc{gnaw}$\oplus$1 in generative mode, using only one value for the nodes $NEG$, $MCS$, $MIV$, $MT$, $ML$ and $ADJ$, the number of generated sentences is approximately 157 000. By adding another value to the $ADJ$ node, the number of realizations increases to approximately 630K. This means that the new 473K sentences differ only in the adjective. This situation presents a significant bias and could cause overfitting of the nodes. 
To solve the above problem, we created an algorithm for introducing symbolic labels during the generation process, which will be replaced stochastically by actual values a posteriori. This stage is described in detail in the section \ref{sec:refinamiento}.

To make the artificial corpus more realistic, we implemented a paragraph segmentation algorithm.
Instead of having a single document of $P$ (relatively small) sentences, we regrouped sets with a variable number of sentences (characterized from the $\pi$-{\sc yalli} corpus) by introducing end-of-paragraph tags. This produces an artificial text that is closer to the reality of authentic Nawatl documents.
This stage will also be described in section \ref{sec:refinamiento}

\subsection{Semantic filters}
\label{sec:filtros}

The artificial corpora generated with both micro-grammars should contain useful sentences (at the grammatical, lexical, and semantic levels) in order to train language models effectively. To retain only the best artificial sentences (those that are both grammatically correct and semantically acceptable), the following could be introduced:

\begin{itemize}
\item a filter for approximating semantics using supervised classification and self-learning methods;

\item a filter based on the association between verbs and animate/inanimate nouns.
\end{itemize}

The semantic filter based on supervised learning, although very interesting, is computationally demanding and will be implemented in future work. The second filter was effectively implemented in this work.
The filter based on the characteristics of animate and inanimate terms allows a significant subset of sentences that lack easily intelligible semantics to be eliminated at low computational cost. 
This semantic filter associates nouns with animate characteristics (e.g., animals or people) with verbs that conventionally express actions appropriate for these nouns (e.g., walk, eat, fly, etc.), as opposed to characteristics of inanimate nouns (objects, artifacts, phenomena, concepts) and their corresponding verbs.
Some examples of the filter can be seen in the Appendix. \ref{sec:appendix1}. 

This approach allows us to expand the $\pi$-\textsc{yalli} corpus with an additional number of realistic sentences that also adhere to plausible Nawatl grammar.
In summary, based on micro-grammars $\mu$\textsc{gnaw}$\oplus$(0,1), sets of $\mathfrak{F}^{0,1}$ sentences are generated.
These sets, once properly filtered, contain respectively $\mathfrak{F}^{0,1}*$ sentences, where $\mathfrak{F}^{0,1} > \mathfrak{F}^{0,1}*$.
This procedure allows us to retain only a diverse set of grammatically and semantically acceptable sentences, which constitute an artificial and usable corpus of sentences in Nawatl.
In Appendix $2$ we present some examples of sentences produced by both micro-grammars.

$\mu$\textsc{gnaw}$\oplus$0 generates
$\mathfrak{F}^{0} \approx 1\times10^6$ unfiltered sentences, or $\mathfrak{F}^0*=807,093$ filtered sentences.
On the other hand, $\mu$\textsc{gnaw}$\oplus$1 generates $\mathfrak{F}^{1}\approx 1.18\times 10^{12}$ unfiltered phrases, which is a very large number. Applying a semantic filter is unrealistic.
For this reason, we use another sentence filtering strategy based on symbolic tags that limit this combinatorial explosion (see Section \ref{sec:refinamiento}).
The number of sentences produced by grammars can be calculated using the values in tables \ref{TG0} and \ref{TG1} (Knowledge Base in Appendix 4) for terminal nodes:

\begin{table}[h!]
\centering
\resizebox{0.62\textwidth}{!}{%
\begin{tabular}{c|c|c|c|c|c|c|c|c|c}
 \toprule
  $N$ & $v$& $ADV_Q$ & $POS$ & $ART$ & $ADV_T$ & $ADJ$ & $PP$ & $NEG$ & $\mathfrak{F}^0$ \\ \hline 
 26 & 16 & 5 & 3 & 3 & 7 & 3 & 3 & 3 & $1,014,993$ \\  
 \bottomrule 
\end{tabular}
}
\caption{Knowledge base for $\mu$\textsc{gnaw}$\oplus$0.}
\label{TG0}
\end{table}

\begin{table}[h!]
\centering
\resizebox{0.72\textwidth}{!}{%
\begin{tabular}{c|c|c|c|c|c|c|c|c|c|c|c}
 \toprule
  $NEG$ & $MT$ & $MIV$ & $MV$ & $MO$ & $v$ & $MCS$ & $ADJ$ & $POS$ & $n$ & $ML$ & $\mathfrak{F}^1$ \\ \hline
 11 & 11 & 4 & 1 & 1 & 27 &  4 & 10 & 4 & 42 & 8 & $1.18\times 10^{12}$ \\  
 \bottomrule 
\end{tabular}
}
\caption{Knowledge base for $\mu$\textsc{gnaw}$\oplus$1 (Only for the \textbf{VSO} structures).}
\label{TG1}
\end{table}

\subsection{Post-processing of $\mu${\sc gnaw}$\oplus$1}
\label{sec:refinamiento}

\begin{figure}[h!]
\centering
\includegraphics[width=0.8\linewidth]{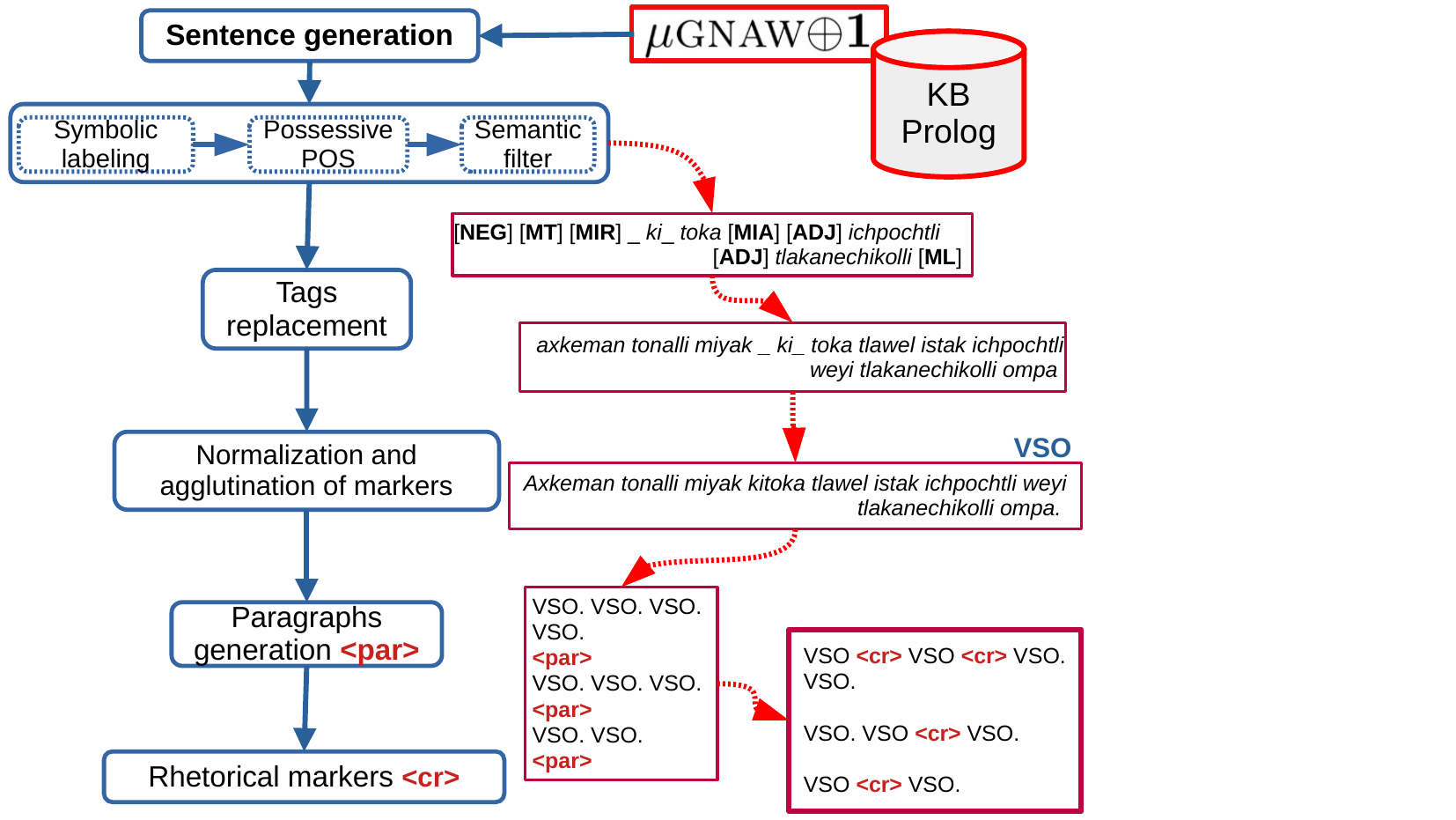}
  \caption{Post-processing for $\mu${\sc gnaw}$\oplus$1.}
  \label{fig:refinamiento}
\end{figure}

This section details the processes of normalizing sentences (cleaning, capitalization, etc.), introducing symbolic tags to control the combinatorial explosion of the grammar and to promote diversity in its realizations. 
It also describes the processes of segmentation into paragraphs and the stochastic introduction of rhetorical connectors with the aim of producing realistic artificial text for our purposes. 
These stages are illustrated in Figure~\ref{fig:refinamiento} and described below.

\begin{description}
\item [Symbolic tagging.]
This labeling consists of replacing the values of the nodes $NEG$, $MCS$, $MIV$, $MT$, $ML$, and $ADJ$ with their respective tags {\sc [neg], [mcs], [miv], [mt], [ml]} and \textsc{[adj]}. 
These tags are then replaced with values from the terminal nodes, respecting the distribution probability in the authentic corpus. This gives the artificial corpus more realistic characteristics.

\item [Possessive management $POS$.]
Phrase generation transforms nouns when they are possessed by a $POS$ particle. For example, if \textit{ichpochtli} (\textbf{lady} in English) is used, and you want to say: \textbf{my lady}, the absolutive \textit{tli} must be removed. Thus, \textbf{my lady} is written: \textit{no} (\textbf{my}) + \textit{ichpoch(tli)} = \textit{noichpoch}. 
Appendix \ref{sec:appendix4} shows the possessive form of all nouns. They are distinguished by the particle enclosed in parentheses, which must be removed as in the previous example.

\item [Paragraph generation.] The paragraph tagger applied to the corpus $\pi$-{\sc yalli} generates 8,767 paragraphs from its 364,233 sentences. 
On the other hand, $\mu${\sc gnaw}$\oplus$1 produces 1,956,750 sentences in a single long paragraph. 
This paragraph was segmented respecting the proportion of paragraphs found in $\pi$-{\sc yalli}.
This yields 41,220 artificial “paragraphs” that replicate the number of authentic paragraphs. This allows models to learn from more sentences that are transformed into paragraphs, rather than learning from a single long sentence.

\item [Introduction of rhetorical connectors.] A paragraph can consist of one or more $VSO$ structures. Each paragraph is analyzed and, given the probability of a rhetorical connector, the “.” separating two $VSO$ structures is replaced by a rhetorical connector.

\item [Text normalization.] This process requires the following steps: removing repeated phrases, concatenating words, normalizing spaces, capitalizing the first letter, and inserting a period.

\end{description}

Table \ref{tab:stat_grammaires} shows the basic statistics for both grammars, used in generative mode.

\begin{table}[h!]
\centering
\footnotesize
\resizebox{0.42\textwidth}{!}{%
 \begin{tabular}{c|r|r|r}
  \toprule
  \bf Grammar & \bf Tokens & \bf Sentences & \bf Paragraphs  \\ \hline
  $\mu${\sc gnaw}$\oplus$0 & $\approx$ 4.6M & $\approx$ 1M  & 714  \\ \hline
  $\mu${\sc gnaw}$\oplus$1 & $\approx$ 6M & $\approx$ 1.9M & 41 220 \\
  \bottomrule
\end{tabular}
}
\caption{\small 
 Statistics on artificial corpora $\mu${\sc gnaw}$\oplus$($\bullet$). Central Nawatl variant.}
  \label{tab:stat_grammaires}

\end{table}

\subsection{Merging of artificial sentences and the Nawatl corpus $\pi$-{\sc yalli}}
\label{sec:corpus}

Below we present some characteristics of the “authentic” Nawatl $\pi$-{\sc yalli} corpus\footnote{Texts written by humans, as opposed to artificial texts generated by algorithms.} \cite{piyalliTALN}, which we have used in our semantic similarity experiments.
The corpus is heterogeneous in terms of linguistic categories and variants of Nawatl (especially from Mexico and Nawatl from El Salvador, see Table~\ref{tab:statcorpus}), and contains a relatively small number of tokens ($\approx$~6.63M) and sentences ($\approx$~364K).
This makes it useful for training classical language models (vector models, TF.IDF, etc.) or static models (Word2Vec, FastText, or Glove embeddings), but unsuitable for training contextualized language models (using BERT-type transformers).
In fact, it has been reported that contextual LLMs require between 10 million and 100 million tokens to obtain stable embeddings.~\cite{micheli2020importancepretrainingdatavolume}.

\begin{table}[h!]
\centering
\footnotesize
\resizebox{0.9\textwidth}{!}{%
 \begin{tabular}{c|r|r|r|r}
  \toprule
  \bf Topics & \bf Docs & \bf Variants & \bf Tokens & \bf \% \\ \hline
  AGR & 3 & cen(2),hua(1) & 7 828 & 0.12 \\ \hline
  
  COS & 6 & ver(1),hua(2), pue(1),cla(1),gue(1) & 53 408 & 0.81\\ \hline
  
  ECO & 1 & cen(1) & 16 777&0.25 \\ \hline

  EDU & 98 & cen(65),hua(9), pip(7),pue(9),gue(6) & 502 392 & 7.58\\ \hline

  HIS & 56 & cla(47),cen(7), pue(1) & 705 790 & 10.64\\ \hline
  
  LEG & 26 & cen(9),cla(3), pue(4),hid(2),oax(1), hua(5),gue(1),mic(1) & 352 563 & 5.32 \\ \hline
  
  LIN & 13 & cen(5),hua(5), cla(1),pip(1) & 402 364 & 6.07 \\ \hline

  LIT & 138 & mix(41),cen(38),pue(34), gue(12),hid(4),tla(5), mor(4) & 1'018 669 & 15.37 \\ \hline

  MED & 4 & cen(2),hua(2) & 14 250&0.23 \\ \hline
  MUS & 5 & cen(5) & 4 306 &0.07 \\ \hline
  PHR & 49 & cen(42), mix(7) & 9 259 &0.15 \\ \hline
  POE & 12 & cen(9),gue(1),mix(2) & 6 604 &0.10 \\ \hline
  POL & 3 & mor(2),cen(1)& 1 800 &0.03 \\ \hline

  REL & 31 & cla(15),cen(4), pue(3), gue(3), oax(1),hua(3),mix(2) & 3'311 474 & 49.95\\ \hline

  TEC & 3 & cen(1),pue(1),hua(1) & 27 838 & 0,42 \\ \hline
  WIK & 4 298& mezcla de variantes &194 292& 2,93\\ \hline
 
  \multirow{1}{*}{\bf Total} & \multirow{1}{*}{\bf 4 746} &  cen(190), cla(67), pue(53), hua(29), hid(6),oax(2), mor(6),pip(8) & \multirow{1}{*}{\bf 6.629M}& \multirow{1}{*}{\bf 100.00} \\  \hline
  \multicolumn{5}{c}{\bf 364K Sentences} \\ 
  \bottomrule
\end{tabular}
}
\caption{\small 
 Document statistics for corpus $\pi$-\textsc{yalli} v1.10. Linguistic variants: cen: central, cla: classical, pue: Puebla, gue: Guerrero, oax: Oaxaca, mor: Morelos, hua: Huasteca, hid: Hidalgo, pip: nawat El Salvador; in parentheses, the number of documents per variety. In several cases there may be a mixture (mix).
 The categories are as follows:
 AGR: Agricultura; COS: Cosmovision; ECO: Economy; EDU: Education; HIS: History; LEG: Legal Documents; LIN: Linguistics; LIT: Literature; MED: Medicine; MUS: Music; PHR: General sentences; POE: Poetry; REL: Religion; TEC: Science and Technology; WIK: Wikipedia.
  \label{tab:statcorpus}
 }
\end{table}

The next step is to combine this “authentic” corpus with an artificially generated corpus. We believe that this combination may be beneficial in the learning of static embeddings and for language models (possibly lightweight LLMs \cite{liu-etal-2024-llmembed}). Depending on the available tokens (authentic or artificial), we have focused on training static embedding representations, because they require less data for learning purposes.
After combining the $\pi$-{\sc yalli} corpora with the artificial sentences (filtered and normalized) produced by the micro-grammars $\mu$\textsc{gnaw}$\oplus$(0,1), the corpus thus created undergoes an appropriate spelling unification process in order to standardize the use of characters in the different variants of Nawatl\footnote{For example, changing c to k, hu to w, eliminating double consonants, using lowercase letters, avoiding accents, etc.}.
Indeed, it has been reported that the unification of lexical variants produces more stable results by reducing token variability\cite{MICAI-piyalli-unigraph}.
The new corpus, $\pi$-{\sc yall-ia}$\oplus$(0,1), was used in FastText static embedding learning with the aim of improving its performance in the task of semantic similarity at the sentence level.

\section{Experiments}
\label{sec:experiments}

In this section, we present a protocol for evaluating our proposal. Semantic similarity assessment is a classic task in NLP. It involves evaluating various models (statistical models, neural networks, etc.) using standardized evaluation protocols \cite{francis-landau-etal-2016-capturing}.
We focus our attention on the task of semantic similarity between a reference sentence and a set of candidate sentences.
This results in a ranking of candidate sentences that can be directly compared to a ranking generated by humans, via a statistical estimator.
This is the evaluation protocol used in \cite{piyalliTALN}, which we adopt here by training embeddings to evaluate the impact of learning based on existing corpora and artificial corpora.

\begin{figure}[h!]
\centering
\includegraphics[width=0.65\linewidth]{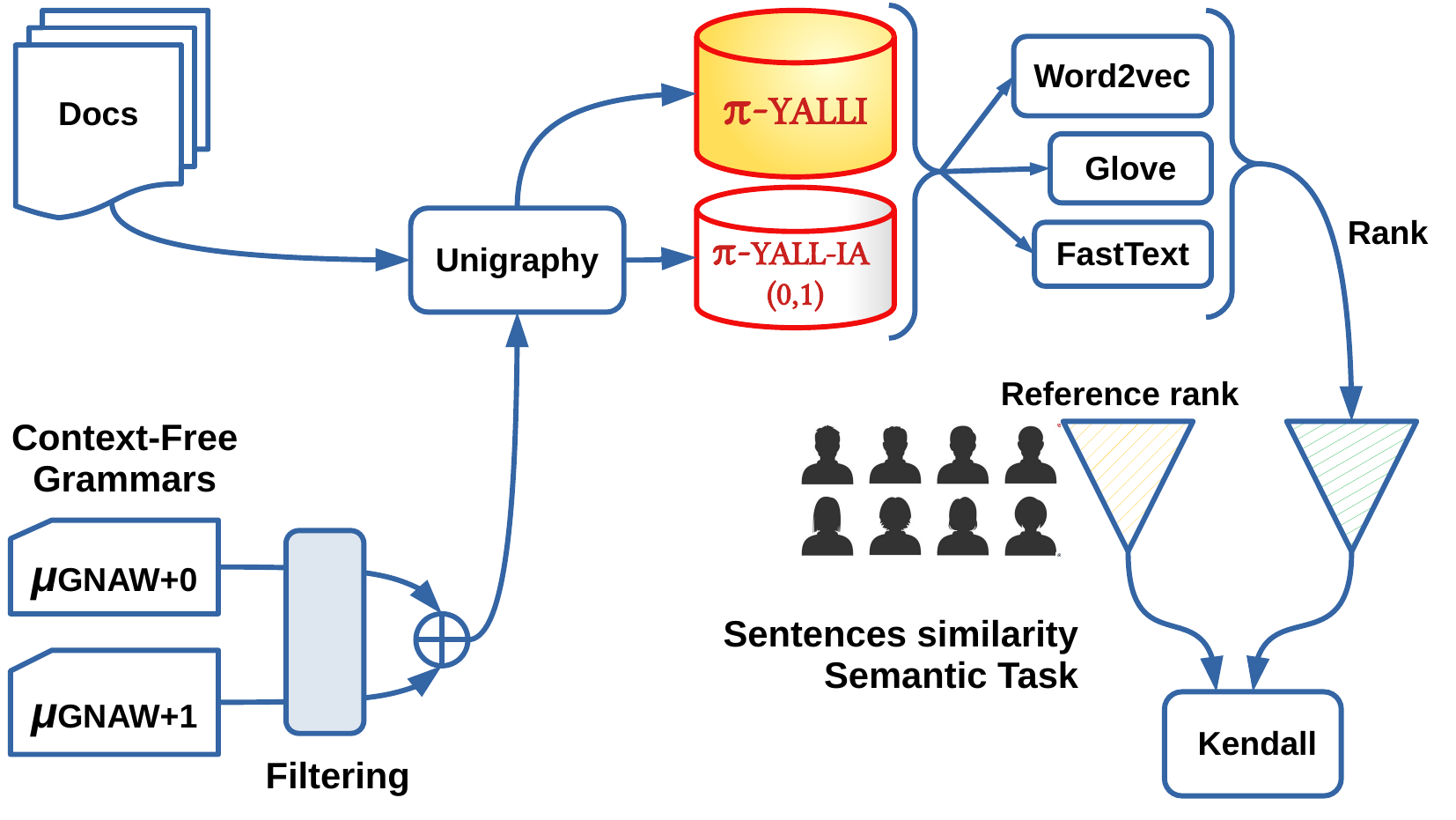}
  \caption{Protocol for evaluating the Semantic Sentences Similarity task.}
  \label{fig:flow}
\end{figure}

The Figure~\ref{fig:flow} illustrates the  experimental flow, from corpus construction (derived from authentic documents or formal grammars), spelling unification, to semantic task evaluation through a reference ranking produced by human annotators, compared with the model's ranking.
We did not expect to achieve high performance in the semantic similarity task using only the artificial corpora generated by the micro-grammars $\mu$\textsc{gnaw}$\oplus(\bullet)$.
Rather, our goal is to enrich the $\pi$-\textsc{yalli} corpus with new artificial datasets containing frequently used grammatical structures.
Another of our hypotheses is that the corpus $\mu$\textsc{gnaw}$\oplus$1 —which better models the Nawatl grammar— should contribute to increasing the performance of embedding-based models.

The extended corpora $\pi$-{\sc yall-ia}$\oplus(\bullet)$ were used to train the FastText algorithm \cite{bojanowski-etal-2017-enriching} from scratch.
The resulting embeddings were subsequently used in the task of semantic similarity between sentences.
A total of $R=30$ reference sentences were written in Spanish by an annotator. These $R$ sentences were divided into 6 blocks of 5 sentences each. For each block, an annotator was assigned, who produced $C=5$ candidate sentences with varying degrees of semantic similarity to the reference sentence.
The 180 sentences (30 reference sentences and 150 candidate sentences) were translated into the Central Nawatl variant by a bilingual Nawatl-Spanish speaker.
The semantic similarity task between sentences used is as follows: given the set of reference sentences $r_i$, $i=1,2,...,R$ (without context), where each reference sentence is associated with a list $\mathcal{L}_i$ of 5 candidate sentences, the objective is to order the lists $\mathcal{L}_i$ according to their semantic similarity to the corresponding reference sentence.
The basic statistics for this task are presented in Table \ref{tab:stats_piyalli}.

\begin{table}[h!]
\centering
\begin{center}
\resizebox{0.5\textwidth}{!}{%
 \begin{tabular}{r|c|c|c|c}
 \toprule
  & \bf Sentences & \bf Tokens & \bf Types &\bf Tokens \\ 
  &  &  &  &\bf per sentence \\ \hline
  \bf References & 30 &246 & 189 & 8.20 \\ \hline
  \bf Candidates & 150 &1026 & 599  & 6.84 \\ \hline
  \bf Total & 180 &1272 & 788  &   \\ \bottomrule  
 \end{tabular}
 }
\end{center}
\caption{Basic statistics of the semantic sentences evaluation task.}
\label{tab:stats_piyalli}
\end{table}

Appendix \ref{sec:appendix4} shows two examples of reference–candidate sentence blocks that are included in the reference ranking ($R_R$).
The reference ranking $R_R$ is produced by a consensus among five human annotators. The objective of the task is to measure how close the rankings $R_M$ generated by various language models (static and contextualized) are to the reference ranking $R_R$.
The measure of similarity between rankings is estimated using Kendall's $\tau(R_R, R_M)$, a nonparametric measure of correlation that evaluates the ordinal association between two variables, i.e., the degree of agreement between two rankings $x$, $y$ (with tie correction):
\begin{equation}
\tau = \frac{C - D}{\sqrt{(C + D + T_x)(C + D + T_y)}}
\label{eq:kendall_corr}
\end{equation}
where $C$ is the number of matching pairs,
$D$ the number of discordant pairs,
$T_x$ the number of tied pairs in $x$ and
$T_y$ the number of tied pairs in $y$ \cite{665905b2-6123-3642-832e-05dbc1f48979}.

\subsection{Non contextual embeddings}

While it is true that transformers have demonstrated superiority in NLP tasks that use computationally rich languages, the situation changes when processing $\pi$-languages.
Static embeddings can be generated from scratch, are quick to produce, and can be used with small corpora containing a certain number of words that appear rarely. 
Examples of these algorithms are Word2Vec \cite{Mikolov2013distributed}, FastText \cite{bojanowski-etal-2017-enriching}, and Glove \cite{glove}. 
Also, we found three sets of FastText and Word2Vec embeddings that are pre-trained and available on Nawatl:
\begin{itemize}
    \item FastText pre-trained on Common Crawl corpora\footnote{https://commoncrawl.org/}; 
    \item FastText pre-trained on Wikipedia in Nawatl\footnote{FastText has been trained on 157 languages worldwide. See the website: \url{https://fasttext.cc/docs/en/crawl-vectors.html}};
    \item Word2Vec in Skip-gram mode pre-trained by John Snow Labs\footnote{See website: \url{https://sparknlp.org/2022/03/16/w2v_cc_300d_nah_3_0.html}}.
\end{itemize}

The embeddings were tested with and without the unification process \cite{MICAI-piyalli-unigraph} so as not to penalize them with respect to our own embeddings.
These pre-trained models have 300 dimensions.
We decided to compare our results on these sets of pre-trained embeddings with FastText embeddings trained on our corpora.

In this article, the implementation of FastText\footnote{The Python package used was: \url{https://radimrehurek.com/gensim/models/fasttext.html}} was trained using the following hyperparameters:
 Iterations: 20; Window: 5; Mode: Skip-Gram; Dimensions: 300

\cite{Mikolov2012Context} discovered that Skip-Gram works well with small corpora and can better represent less frequent words.
That is why we decided to use this implementation with the Nawatl texts.

\subsection{LLM models used}

In our experiments, we used the following five LLM models via APIs\footnote{
Access to the LLMs via API was carried out on June 9, 2025, from 4 p.m. to 8 p.m. GMT. Testing with the interactive models (Copilot, Grok, and Claude) was carried out on June 9 and 10, 2025.}: ChatGPT-4 mini API; Gemini-2.5-flash-preview-05-20 API\footnote{\url{https://deepmind.google/models/gemini/flash/}}; DeepSeek-V3-0324 API\footnote{\url{https://api-docs.deepseek.com/news/news250325}}; Llama-3.1-70B-Instruct API\footnote{\url{https://huggingface.co/meta-llama/Llama-3.1-70B-Instruct}}
 and Mistral-large-latest~API\footnote{\url{https://mistral.ai/news/mistral-large}}.
In interactive mode we used three  models: Copilot (\url{https://www.microsoft365.com}), Grok 3 (\url{https://grok.com}) and Claude 3.7 (\url{https://claude.ai}).
\color{black}
The prompt used was the following:\footnote{
The prompt was written in French: <<{ \it
\noindent \`A partir de la phrase Nawatl: ``[référence.]'' triez par ordre sémantique, de la plus proche à la plus lointaine, les cinq phrases suivantes: ``[candidate$_1$.]''; ``[candidate$_2$.]''; ``[candi\-date$_3$.]''; ``[candidate$_4$.]''; ``[candi\-date$_5$.]''.  Ne donnez pas des rankings de phrases autres que celles proposées. }>>}
{ \it
\noindent Given the Nawatl sentence ``[reference.]'' rank the following five sentences semantically, from the closest to the furthest in meaning from the original sentence:
``[candidate$_1$.]''; ``[candidate$_2$.]''; ``[candidate$_3$.]''; ``[candidate$_4$.]''; ``[candidate$_5$.]''.
Do not give ranking of other sentences than those provided.} The references and candidates were written in Central Nawatl.

\color{black}

To avoid bias on our evaluation, we used a cross-validation protocol (Leave-one-out) \cite{moss-etal-2018-using}, which avoids possible overfitting biases. We performed five runs at each evaluation point.

\subsection{Results and discussion}

\begin{figure}[h!]
  \centering
  \includegraphics[width=0.7\linewidth]{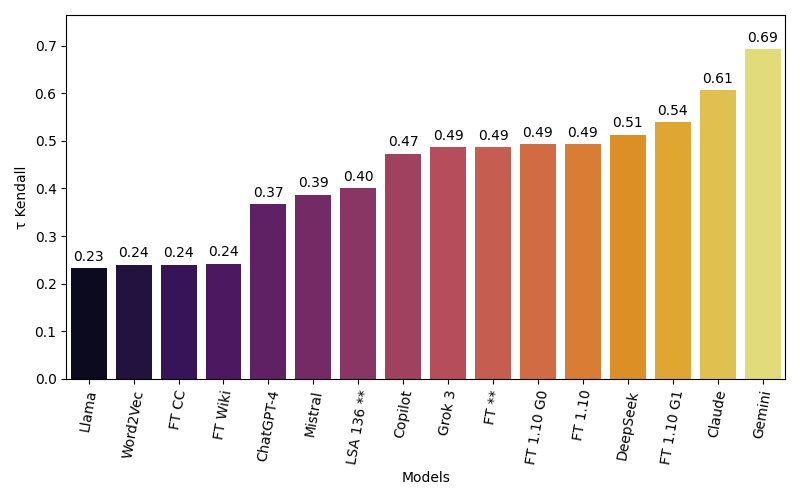}
    \caption{Kendall's $\tau$: LLMs, FastText/Common Crawl-Wikipedia (FT CC / FT Wiki) and Word2Vec vs FastText/$\pi$-{\sc yall-ia} Skip-gram in 300 dimensions, without empty words: \textit{iwan, in, tlen} \& \textit{ipan}. Maximum values obtained in {\it Leave-3-out} evaluation are shown. ** From \cite{MICAI-piyalli-unigraph}}
    \label{fig:barras}
\end{figure}

Figure~\ref{fig:barras} summarizes all our results.
The best result was obtained using FastText on the $\pi$-\textsc{yall}-\textsc{ia}$\oplus$1 corpus, which achieved a maximum Kendall $\tau$ = {\bf 0.540} (FT 1.10 G1).
This places the FastText algorithm, trained on the expanded corpus (authentic sentences and artificial data), in third place in terms of performance, surpassed only by the Gemini 2.5 and Claude 3.7 LLMs.

\color{black}
  
Indeed, our hypothesis regarding the expansion of the Nawatl corpus proved more reliable with grammar $\mu$\textsc{gnaw}$\oplus$1.
This suggests that a more accurate modeling of Nawatl grammar does help to capture the deeper structures of the language— such as agglutination, polysynthesis, and the central role of the verb— resulting in a (slight, yet directionally meaningful) improvement in performance on the semantic similarity task involved.

\section{Conclusions}
\label{sec:conclusions_future}

The generative use of micro-grammars $\mu$\textsc{gnaw}$\oplus$(0,1) allows for the creation of a large number $\mathfrak{F}^{0,1}$ of sentences, which can be considered as a corpus —artificial and very extensive— but which presents redundancy and poor semantics.
A semantic filtering process was applied in order to reduce the universe to $\mathfrak{F}^{0,1}*$ sentences, decrease redundancy, and achieve slightly more plausible semantics.
Subsequently, the artificial corpus was concatenated with the authentic corpus. $\pi$-\textsc{yall}-\textsc{ia}$\oplus$1, a corpus resulting from the union of $\mu$\textsc{gnaw}$\oplus$1 (closer to the structures of the Nawatl language) and $\pi$-\textsc{yalli}, enabled efficient learning using a representation of static embeddings.
Using $\pi$-\textsc{yall}-\textsc{ia}$\oplus$1, the FastText algorithm improved its performance in the task of evaluating semantic similarity between sentences, increasing  the Kendall $\tau_{\textsc{max}}$ from 0.493 to 0.540 (+9\% increase).
It should be noted that in this task, we also compared FastText and Word2Vec embeddings pre-trained on other corpora, and their results are disappointing ($\tau \approx$ 0.242) compared to FastText trained on $\pi$-{\sc yalli}.
Using a formal Nawatl micro-grammar in generative mode seems to favor learning FastText algorithms because the embeddings better capture the structure of the Nawatl language. 

In future work on the grammar $\mu$\textsc{gnaw}$\oplus1$, we will seek to increase the number of elements n and v, include additional grammatical persons, base 2 and 3 verbs, plurals, more rhetorical connectors, markers, and improve the semantic filter (to avoid combinatorial explosion effects).
Another idea to explore is the introduction of Probabilistic Context-Free Grammars (PCFG), where nodes can have associated probabilities. This type of grammar\footnote{\url{https://en.wikipedia.org/wiki/Probabilistic_context-free_grammar}}
 \cite{CFGprob} is particularly useful when filtering semantic/non-semantic sentences using an external classifiers. This classifier begins with manual labeling and then produces automatic labeling of a subset of phrases. This classification could be used to modify the parameters (in this case, the probabilities) of the generative grammar.

\color{black}

\section*{Acknowledgments}

This research work is funded by an Intermedius PhD thesis grant (UA, France), and partially funded by the Université d'Avignon, Laboratoire Informatique d'Avignon (LIA), and the Agorantic research program (France).
M.F.-S. and L.Q.T are also grateful for the kind hospitality of the Laboratoire Informatique d'Avignon (LIA) and Intermedius School, Université d'Avignon, where they participated in conducting of this research.

\bibliographystyle{apalike}
\bibliography{references}

\section*{Appendix 1: Filtering animated/inanimate elements}
\label{sec:appendix1}

This filter uses tags for nouns and verbs that can be categorized as \textit{animate} or \textit{inanimate} in the intuitive sense of the term.
The filter is implemented as follows: a sentence is constructed using only nouns associated with verbs of the same type ($n$ animate/$v$ animate or $n$ inanimate/$v$ inanimate). Some examples of appropriate verbs and nouns are shown below.

\begin{description}
    \item[Nouns:]~
    \begin{itemize}
    \item \textit{nantzin} (mother):  \textbf{animate}
    \item \textit{tiotzin} (god): \textbf{animate / inanimate}
    \item \textit{tatzin} (father): \textbf{animate}
    \item \textit{tlahtolli} (speech):  \textbf{inanimate}
    \item \textit{mapachin} (raccoon): \textbf{animate}
    \item \textit{kuawtli} (eagle): \textbf{animate}
    \item \textit{momachtiani} (student): \textbf{animate}
\end{itemize}

\item [Verbs:]~
\begin{itemize}
    \item \textit{mawisowa} (admire): \textbf{animate}
    \item \textit{neki} (want): \textbf{animate}
    \item \textit{pia} (to have): \textbf{animate / inanimate}
    \item \textit{itta} (to see): \textbf{animate}
    \item \textit{chihua} (to do): \textbf{animate}
    \item \textit{toka} (bury): \textbf{animate / inanimate}
\end{itemize}
\end{description}

\color{black}

The rule states that if the labels match, the sentence can be generated. That is, the nouns father (\textit{tatzin}), mother (\textit{nantzin}) or raccoon (\textit{mapachin}) can be generated with the verbs want (\textit{neki}) or see (\textit{itta}), but for the noun speech (\textit{tlahtolli}) is not the case.

\section*{Appendix 2: Examples of artificial paragraphs generated by $\mu$\textsc{gnaw}$\oplus$1}
\label{sec:appendix2}
The text generated was subsequently processed by the \textit{unigraphic} system \cite{MICAI-piyalli-unigraph}:

\begin{description}
    \item[1] {\it Kimachtia itoto tonatih. Kimachtia itoto sentilistli. Kimachtia itoto momachtiani. Kimachtia itoto temachtiani. Tlen Kimachtia itoto posoli.}

    \item[2]{\it Kitoka mokoltsin noxochih. Kitoka mokoltsin nosiwah. Kitoka mokoltsin notatsin. Kitoka mokoltsin nonantsin. Kitoka mokoltsin nototo. Iwan Kitoka mokoltsin nokoyo. Kitoka mokoltsin notateh. Inic Kitoka mokoltsin ichpoch. Kitoka mokoltsin itlakanechikol. Iwan Kitoka mokoltsin imapach. Kitoka mokoltsin iyolka. Kitoka mokoltsin itelpoch. Kitoka mokoltsin ikuaw. Ka kitoka mokoltsin ilamah.
    [~$\cdots$ ]     
Kitoka mokoltsin imapach. Kitoka mokoltsin iyolka. Kitoka mokoltsin itelpoch ahko. Kitoka mokoltsin ikuaw. Kitoka mokoltsin ilamah. Kitoka mokoltsin iweweh. Kitoka mokoltsin isiwamich. Kitoka mokoltsin itlaka.}
    
\end{description}

\color{black}

\section*{Appendix 3: Semantic Similarity Sentences Task}
\label{sec:appendix3}

This appendix shows two examples of the semantic similarity classification task. Candidate phrases are sorted from highest to lowest semantic similarity with respect to their reference.
The corpus and programs for this task are available on the  website:
\url{https://demo-lia.univ-avignon.fr/pi-yalli}

\noindent 
~\\
\textsc{Reference$_3$:} 

\textbf{\textit{Tonatih kipalewia moskaltiah kilitl iwan xokokuawitl.}} 
{ \rm The sun serves to make green plants and fruit trees grow.}\\

\noindent \textsc{Candidats:}
  \begin{enumerate}
  \item Moneki xiwimeh kiseliah tlawili, atl pampa moskaltiskeh. / \textit{Vegetation needs to receive light and water to grow.}
  \item Kuawmeh kinmanawiah xiwimeh iwikpa tonalmitl. / \textit{Trees protect plants from the sun's rays.}
  \item Amo nechpaktia nikilikua ipanpa ok achi nechpaktia tlakilotl. /
  \textit{I don't like eating vegetables because I prefer fruit.}
  \item Tonaltlawili ekolohika ipanpa xoxowik tlawili, no ihkin atlawili. /
  \textit{Solar energy is environmentally friendly because it is a green energy source, like water energy.
  }
  \item Ihkuak se kiselia miak tonalmitl weli kinextia kualokatl. / 
  \textit{Receiving too much sunlight can cause cancer.}
\end{enumerate}~

\noindent \textsc{Reference$_4$:} 

\textbf{\textit{In altepetl se kitlalpachowa, moketsa iwikpa ahiektli tekiwahkayotl.}}

{\rm The oppressed people rebel against the tyranny of their rulers.}\\

\noindent \textsc{Candidats:}
\begin{enumerate}
  \item Ahmelawak tekiwah nochipa kipehpewaltia in alteperebolosion / 
  \textit{A dictator always causes a popular revolution.}
  \item Ihkuak se kitlalpachowa se tlakanechikoli, yeh kineki moketsas tewikpa. /   \textit{An oppressed people seeks to rebel.}
  \item Inik tlakanechikoli moketsa tewik, tekiwahkeh motekipachowah. /
  \textit{The rebellion of the people worries their rulers.}
  \item Tlakanechikolmeh kipiyah tekiwahkayotl tlen moneki. / \textit{People get the governments they deserve.}
  \item In repoblikah se tlamantli tekiwahkayotl. /  \textit{A republic is a system of government.}
\end{enumerate}

\color{black}
\section*{Appendix 4: Knowledge bases (KB) in Prolog for $\mu${\sc gnaw}$\oplus1$}
\label{sec:appendix4}
The following table shows the roots of the 27 nouns used (with the absolutive particle that disappears in parentheses). Similarly, for each of the formal grammars, tables are shown for the verbs, adverbs, adjectives, possessives, and markers used in the Prolog sentence generator.

\begin{table}[h!]
\centering
\footnotesize
\resizebox{0.83\textwidth}{!}{%

\begin{tabular}{|c|c|c|c|c|}
  \hline
  \multicolumn{5}{|c|}{\bf Nouns }\\ \hline
  \bf ichpoch(tli) & \bf tlakanechikol(li) & \bf mapach(in) & \bf yolka(tl) & \bf telpoch(tli) \\
  \it young woman & \it group of people & \it raccoon & \it beast of burden & \it young man \\ \hline
  \bf kuaw(tli) & \bf koltzin & \bf ilamah & \bf weweh & \bf cihuamich(in) \\
  \it eagle & \it grandfather & \it old woman & \it old man & \it mermaid \\ \hline
  \bf tlaka(tl) & \bf tlamatini & \bf tonatih & \bf sentilis(tli) & \bf momachtiani \\ 
  \it man/person & \it wise man/wise woman & \it sun & \it family & \it student \\ \hline
  \bf temachtiani & \bf posol(li) & \bf xochi(tl) & \bf siwa(tl) & \bf tlahtol(li)* \\
  \it teacher & \it pozole & \it flower & \it woman & \it speech/language/narrative \\ \hline
  \bf tatzin & \bf nantzin & \bf tiotzin** & \bf toto(tl) & \bf coyo(tl) \\ 
  \it father & \it mother & \it god & \it bird & \it coyote \\ \hline
  \bf tateh & \bf tonalli & & & \\ 
  \it gentleman & \it by day & & & \\ \hline 
\end{tabular}
}
\caption{Nouns in grammar $\mu$gnaw$\oplus1$. Inanimate nouns are marked with a \textbf{*} and those that are animate/inanimate are marked with \textbf{**}.}
\end{table}


\begin{table}[h!]
\centering
\footnotesize
\resizebox{0.87\textwidth}{!}{%

\begin{tabular}{|c|c|c|c|c|}
\hline
 \multicolumn{5}{|c|}{\bf Verbs}\\ \hline
 \bf toka** & \bf kaki & \bf ahsikamati & \bf itta & \bf chihua \\ 
 \it to bury & \it to listen & \it to understand & \it to see & \it to do \\ \hline
 \bf chiya** & \bf pia** & \bf mati & \bf maka & \bf ixpantilia \\ 
 \it to look & \it to have & \it to feel good & \it to give & \it to show [sth] to [sb] \\ \hline
 \bf machtia & \bf welitta & \bf neki & \bf tlasohtla & \bf paka \\ 
 \it to teach [sth] to [sb] & \it to like [sth]/[sb] & \it to want & \it tolove/appreciate & \it to wash \\ \hline
 \bf paktia & \bf ahawiltia & \bf palewia & \bf tlalpachowa & \bf mawisoitta \\ 
 \it to cheer [sb] up/make happy  & \it to make [[sb] enjoy & \it to assist & \it to subdue & \it to fear with devotion \\ \hline
 \bf mawisowa & \bf kopina & \bf nemilia & \bf elnamiki &  \\ 
 \it to admire & \it to trace/copy & \it to imagine & \it to remember &  \\ \hline \hline

 \multicolumn{5}{|c|}{\bf Rhetorical connector}\\ \hline
 \bf iwan & \bf tlen & \bf ipan & \bf pampa & \bf auh \\
 \it and/with this & \it what & \it in/at & \it because & \it but/so/and \\ \hline
 \bf inic & \bf huel & \bf ihui & \bf ye & \bf ya \\
 \it for this reason & \it well & \it like that & \it already & \it already \\ \hline
 \bf yuhqui & \bf çan & \bf zan & \bf  & \bf  \\ 
 \it in this way/like that & \it only/precisely & \it only/precisely & \it  & \it  \\ \hline \hline

 \multicolumn{5}{|c|}{\bf Temporary marker}\\ \hline
 \bf aman & \bf axkan & \bf yalwa & \bf axcan & \bf axan \\
 \it now/today & \it now/today & \it yesterday & \it now/today & \it now/today \\ \hline
 \bf nama & \bf niman & \bf ic & \bf namatsin & \bf cemicac \\
 \it now/today  & \it after/then & \it when & \it right now & \it always \\ \hline
 \bf quemmanian & \bf & \bf  & \bf  & \bf  \\ 
 \it sometimes & \it & \it  & \it  & \it  \\ \hline \hline

 \multicolumn{5}{|c|}{\bf Place marker}\\ \hline
 \bf ahko & \bf tlakpak & \bf ompa & \bf nepa & \bf oncan \\
 \it up & \it up & \it there & \it further & \it further \\ \hline
 \bf aca & \bf ahuic & \bf can & \bf  & \bf  \\
 \it at the top & \it from one side to the other & \it where & \it  & \it  \\ \hline \hline

 \multicolumn{5}{|c|}{\bf Noun quantity marker}\\ \hline
 \bf miyak & \bf nochi & \bf seki & \bf achi &  \\
 \it a lot & \it all & \it a few & \it suficiente & \\ \hline \hline                       
 \multicolumn{5}{|c|}{\bf Verbal intensity marker}\\ \hline 
 \bf tlawel & \bf wel & \bf miakpa & \bf san &  \\
 \it do a lot & \it do a lot & \it frequently & \it only/precisely & \\ \hline \hline

 \multicolumn{5}{|c|}{\bf Noun-adjectives}\\ \hline 
 \bf weyi & \bf istak & \bf ahwiyak & \bf tepitzin & \bf tomawak \\
 \it big & \it white & \it tasty/aromatic & little/a few & fat \\ \hline
 \bf mawistik & \bf tenyo & \bf chipawak & \bf astatik & \bf melawak \\
 \it amazing & \it famous & \it pure & \it bright white & \it correct \\ \hline \hline

 \multicolumn{5}{|c|}{\bf Negation}\\ \hline 
 \bf amo & \bf axkeman & \bf amitla  & \bf axtle & \bf ate \\ 
 \it no/not & \it never & \it nothing & \it nothing & \it nothing \\ \hline 
 \bf amika & \bf axaka & \bf amo keman  & \bf ah & \bf ax \\ 
 \it nobody & \it no & \it never & \it no/not & \it no/not \\ \hline 
 \bf ka & \bf  & \bf   & \bf  & \bf  \\ 
 \it no & \it  & \it  &  &  \\ \hline \hline

 \multicolumn{5}{|c|}{\bf Possessive}\\ \hline 
 \bf no- & \bf mo- & \bf i- &  &  \\
 \it my & \it your & \it her/his/its &  &  \\ \hline \hline

 \multicolumn{5}{|c|}{\bf Marker subject verb}\\ \hline 
 \bf ni- & \bf ti- & \bf  &  &  \\
 \it i & \it you & \it he/she/this &  &  \\ \hline 
\end{tabular}
}
\caption{Knowledge base for micro-grammar $\mu$gnaw$\oplus1$. Animate/inanimate verbs are marked with a \textbf{**}.}
\end{table}

\end{document}